\def\year{2025}\relax
\documentclass[letterpaper]{article} 
\usepackage{aaai25}  
\usepackage{multirow}  
\usepackage{graphicx}
\usepackage{caption}
\usepackage{graphicx}
\usepackage{float} 
\usepackage{array}
\usepackage{subcaption}
\usepackage{stackengine}

\usepackage{amsmath}
\usepackage{times}  
\usepackage{amsfonts,amssymb}
\usepackage{helvet}  
\usepackage{courier}  
\usepackage[hyphens]{url}  
\usepackage{graphicx} 
\usepackage{natbib}  
\usepackage{caption} 
\DeclareCaptionStyle{ruled}{labelfont=normalfont,labelsep=colon,strut=off} 
\usepackage{booktabs}
\usepackage{algorithm}
\usepackage{algorithmic}
\usepackage{newfloat}
\usepackage{listings}
\floatstyle{ruled}
\newfloat{listing}{tb}{lst}{}
\floatname{listing}{Listing}
\title{Anchor Learning with Potential Cluster Constraints for Multi-view Clustering}

\author{
    Yawei Chen\textsuperscript{\rm 1},
    Huibing Wang\textsuperscript{\rm 1}\thanks{Corresponding Author.},
    Jinjia Peng\textsuperscript{\rm 2},
    Yang Wang\textsuperscript{\rm 3*}
    }
\affiliations {
    \textsuperscript{\rm 1}School of Information Science and Technology, Dalian Martime University, Dalian, China\\
    \textsuperscript{\rm 2}School of Cyber Security and Computer, Hebei University, Baoding, China\\
    \textsuperscript{\rm 3}School of Computer Science and Information Engineering, Hefei University of Technology, Hefei, China \\
    \{cyw,huibing.wang\}@dlmu.edu.cn,
    pengjinjia@hbu.edu.cn,
    yangwang@hfut.edu.cn
}
\usepackage{bibentry}
\begin{document}

\maketitle

\begin{abstract}
Anchor-based multi-view clustering (MVC) has received extensive attention due to its efficient performance. Existing methods only focus on how to dynamically learn anchors from the original data and simultaneously construct anchor graphs describing the relationships between samples and perform clustering, while ignoring the reality of anchors, i.e., high-quality anchors should be generated uniformly from different clusters of data rather than scattered outside the clusters.
To deal with this problem, we propose a noval method termed Anchor Learning with Potential Cluster Constraints for Multi-view Clustering (ALPC) method. Specifically, ALPC first establishes a shared latent semantic module to constrain anchors to be generated from specific clusters, and subsequently, ALPC improves the representativeness and discriminability of anchors by adapting the anchor graph to capture the common clustering center of mass from samples and anchors, respectively.
Finally, ALPC combines anchor learning and graph construction into a unified framework for collaborative learning and mutual optimization to improve the clustering performance. 
Extensive experiments demonstrate the effectiveness of our proposed method compared to some state-of-the-art MVC methods. Our source code is available at {https://github.com/whbdmu/ALPC}.
\end{abstract}

\section{Introduction}
With the advent of the big data generation, there is an increasing diversity and complexity of data, which this data often exists in multiple forms \cite{10.1007/978-3-662-44851-9_41,ijcai2023p398,Wei2021FineGrainedIA}. For example, news can be reported in a variety of formats, including text, images and videos. Due to these multiple sources of data, multi-view learning has attracted extensive attention \cite{10.1609/aaai.v37i7.25960,10539269}. Multi-view clustering is a fundamental task in multi-view learning, which tries to explore complementary information between views, discover the intrinsic structure of clusters, and group information from multiple sources \cite{10203657,10.1145/3474085.3475471,10476709}.

In previous investigations, researchers have contributed a vast number of MVC methods, among which subspace learning has received extensive attention and examination as a particularly compelling research issue. MVC methods based on subspace learning usually learn the self-representation matrix and then perform spectral clustering. However, multi-view subspace clustering usually has high time complexity, ${\cal O}\left( {{n^2}} \right)$ or even ${{\cal O}}\left( {{n^3}} \right)$. The high time complexity comes mainly from the construction of the self-representation matrix and the computation of the spectral embedding in the spectral clustering \cite{10.5555/3666122.3668065,10.1609/aaai.v37i7.26057,10.5555/3060832.3060922}.

In order to address this problem, anchor-based multi-view clustering has been proposed in recent years, which aims to learn representative anchors from multiple samples, and then learn the anchor graph through the relationship between the samples and the anchors and thus perform clustering. Obviously, how to learn high-quality anchors is essential. Existing anchor learning strategies can be categorized into two ways: static selection and dynamic learning.
The static selection strategies include random sampling and k-means, where random sampling aims to randomly select data from all the samples to be used as anchors, while k-means directly utilizes the sample center of mass as an anchor. Although the above methods are simple and effective in the process of selecting anchors, the result of selecting anchors is random, which leads to inferior quality of anchors, thus reducing the clustering performance. In order to improve the quality of anchors, dynamic anchor learning become popular. For example, \cite{9646486} integrates anchor selection and subspace graph construction in a unified framework so that the two processes negotiate with each other to improve the quality of the anchors, avoiding the need to select anchors directly from the original samples. \cite{Sun2021ScalableMS}, on the other hand, adaptively learns the consensus anchor graph of all samples in the projection space to be used for clustering.

Although MVC methods for dynamic anchor learning achieve excellent performance, they only learn anchors dynamically through simple low-rank constraints, which may lead to uneven anchor distribution, i.e., some anchors will be scattered outside the data clusters or some data clusters lack corresponding anchors. That is, these methods ignore the reality of anchors, i.e., high-quality anchors should be naturally generated from different data clusters and share the underlying clustering structure with the original data, rather than being scattered outside the data clusters.

\begin{figure}[htb] 
\centering 
\includegraphics[width=0.48\textwidth]{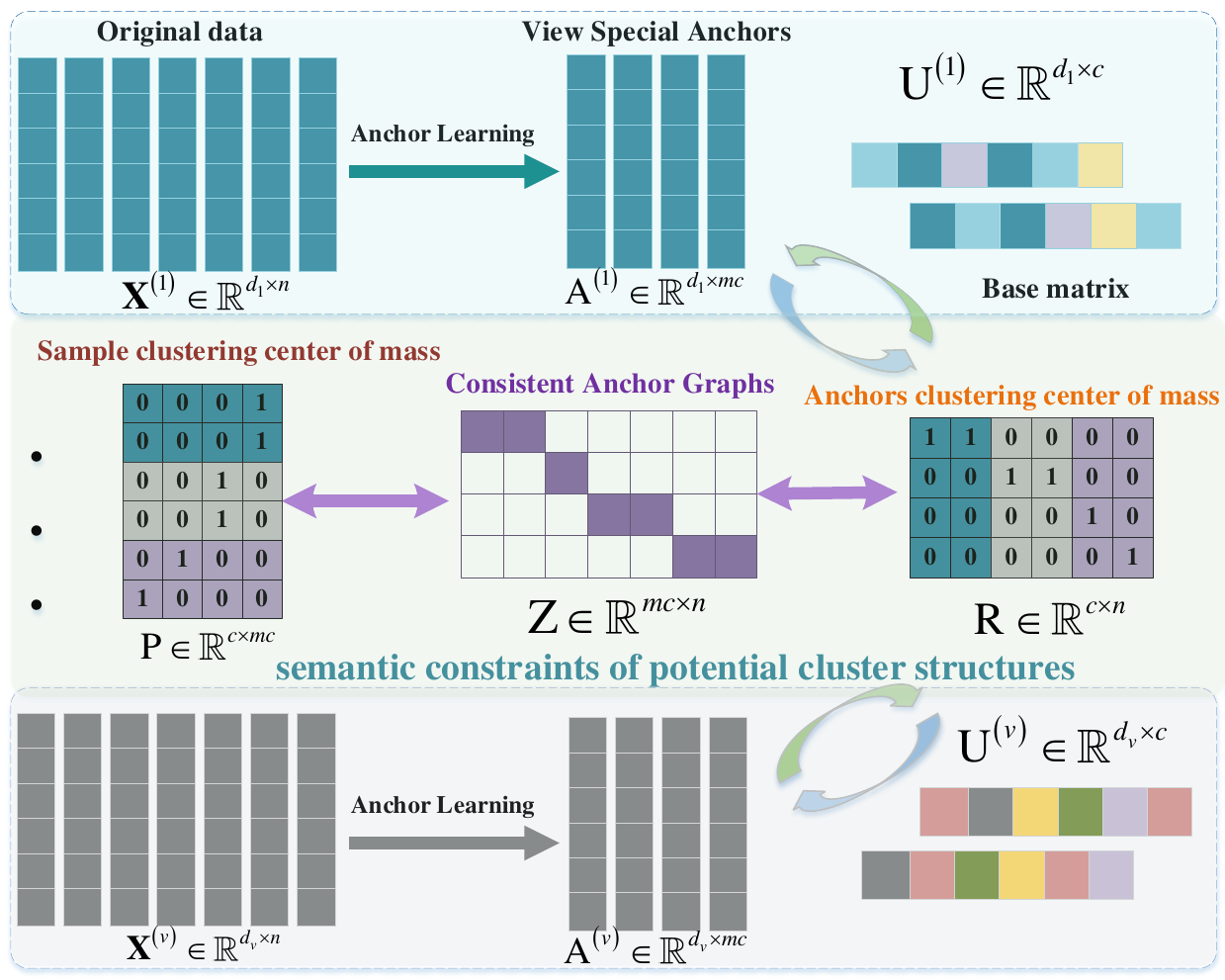} 
\caption{Schematic diagram of the proposed ALPC. Aims to learn the most representative anchors for large multi-view datasets. ALPC unifies anchor learning and consistent anchor graphs into a single model. ALPC adopts the semantic constraint of sharing potential cluster structures with the aim of guiding the anchors to be generated uniformly from different clusters of the original data. At the same time, in order to ensure the representative nature of the generated anchors, ALPC discretizes the anchor graphs to guide the center of mass clustering of the original data and the mass clustering centers of the anchors are aligned, thus constraining both to share the same clustering structure.}  
\label{fig1} 
\end{figure}

In order to solve the above problems, this paper proposes the ALPC method, the framework of which is shown in Fig. 1. Different from the existing multi-view clustering based on anchor graph, firstly, ALPC innovatively proposes the semantic constraint of sharing potential cluster structure, which aims to guide anchors to be generated uniformly in sample clusters, avoiding the situation that anchors are generated outside the clusters or there are no anchors in the sample clusters; secondly, ALPC performs the feature decomposition of the anchor graph to learn the common clustering center of gravity of the samples and anchors, aiming at matching the anchor clustering structure with the original sample clustering structure to match the anchors with a fairer, more discriminative and clearer clustering structure; finally, ALPC unifies anchor selection and potential cluster structure constraints in a single framework so that the two processes negotiate with each other to improve the quality of the anchors, and thus adaptively learns low-rank anchor graphs for clustering.
The main contributions of this paper are totaled as follows when compared to existing MVC methods based anchors:

\begin{itemize}
\item Unlike existing dynamic anchor learning strategies, this paper innovatively proposes to share the semantic constraints of potential cluster structures to adaptively learn representative anchors uniformly from sample clusters.
\item Learning the cluster structure of samples and anchors by fitting the relevance of anchors and learning the cluster structure with differentiation, clarity, and fair anchoring cluster structure by identifying the clustering center of mass of samples and anchors through anchor graphs.
\item Extensive experiments on six benchmark datasets demonstrate the effectiveness of our method compared to state-of-the-art methods.
\end{itemize}

\section{Related Work}
In this section, we will present existing work related to the research in this paper, including methods for multi-view subspace clustering and anchor graph-based multi-view clustering. Table 1 lists the main notations used in this paper.
\begin{table}[!htbp]

\label{tab:tableTab} 
\scalebox{0.899}{
\begin{tabular}{l|l} 
\toprule 
Notations & Descriptions \\
\midrule 
$n$,$l$,$c$         & The number of samples, views and clusters \\
$m$                    &Number of anchors in each cluster \\
${d_v}$     & Feature dimension of the $v$-th view\\
${{\mathbf{X}}^{(v)} \in {\mathbb{R}^{{d_v} \times n}}}$ & Data matrix in the $v$-th view  \\
${{\mathbf{A}}^{(v)} \in {\mathbb{R}^{{d_v} \times mc}}}$     & Anchor matrix of the $v$-th view\\
${{\mathbf{Z}} \in {\mathbb{R}^{{mc} \times n}}}$         & Consensus subspace representation \\
${{\mathbf{U}^{(v)}} \in {\mathbb{R}^{{d_v} \times c}}}$         & Anchored cluster center of mass \\
${{\mathbf{P}} \in {\mathbb{R}^{{c} \times mc}}}$         & Clustering Indicator for anchors \\
${{\mathbf{R}} \in {\mathbb{R}^{{c} \times n}}}$         & Clustering Indicator for original data \\
\bottomrule 
\end{tabular}}
\raggedright
\caption{The Descriptions of Significant Formula Notations} 
\end{table}
 
\subsection{Multi-view Subspace Clustering}
In recent years, multi-view subspace clustering methods have gained extensive attention due to their low-rank representations that can exploit the intrinsic subspace structure of the data to obtain good clustering performance.
Most of these can be described in the following framework:

\begin{equation}
\begin{gathered}
\label{1}
\begin{gathered}
\begin{array}{l}
\mathop {\min }\limits_{{\mathbf{S}^{\left( v \right)}},\mathbf{S}} \sum\limits_{v = 1}^l {\underbrace {\left\| {{{\rm{\mathbf{X}}}^{\left( {{v}} \right)}} - {\mathbf{X}^{\left( {{v}} \right)}}{\mathbf{S}^{\left( v \right)}}} \right\|_F^2}_{Graph{\rm{ }}\quad Construction} + \underbrace {\Omega \left( {{\mathbf{S}^{\left( v \right)}},\mathbf{S}} \right)}_{Fusion}} \\
s.t.diag\left( {{\mathbf{S}^{\left( v \right)}}} \right) = 0,diag\left( \mathbf{S} \right) = 0,\mathbf{S}_i^T\textbf{1} = 1.
\end{array}
\end{gathered} 
\end{gathered} 
\end{equation}
where $\mathbf{S}^{\left( v \right)} \in {\mathbb{R}^{n \times n}}$ denotes the subspace representation of the $v$-th view and $\mathbf{\Omega}$ denotes the uniform consensus regularization term with the aim of obtaining the fusion graph $\mathbf{S}  \in {\mathbb{R}^{n \times n}}$.

Most existing multi-view subspace clustering methods have evolved from the aforementioned framework because they can capture global structures and obtain low-rank graphs. For example, \cite{7298657} proposed inducing the Hilbert Schmidt Independence Criterion (HISC) to learn complementary information from different views, and obtained clustering results by imposing low-rank constraints on the learned affinity graph. \cite{10.5555/3504035.3504492,2021Survey,2022Progressive} divided the subspace representation of each view into a fixed part and a specific part to retain consistency and complementary information between views, making the subspace representation more precise. However, these methods ignore high-order information between affinity graphs, so recently, methods based on low-rank tensor constraints have become popular. These methods stack different views' affinity graphs into a tensor to explore high-order information between views, such as \cite{9813586, 9684399, 10232925}.

The above methods have good performance in clustering. However, these methods cannot avoid constructing a complete ${{n}} \times {{n}}$ graph, which leads to a time complexity of ${\cal O}\left( {{n^3}} \right)$, making them powerless in large-scale data sets and limiting their application in real scenarios.

\subsection{Anchor Based Multi-view Clustering}
In the field of multi-view subspace clustering, anchor graphs have been recognized as an effective means of dealing with large-scale datasets. The principle of the anchor graph approach is to select $m$ representative anchors among $n$ samples and explore the relationship between the anchors and the samples. As a result, the construction of the anchor graph is reduced from $\mathbf{S}  \in {\mathbb{R}^{n \times n}}$ to $\mathbf{Z}  \in {\mathbb{R}^{m \times n}}$, which undoubtedly greatly reduces the computation time as well as storage and maintains the excellent clustering performance.

Recently, a representative of the anchor-based MVC approach, \cite{9646486} proposed a fast and parameter-free multi-view clustering method, i.e., learning anchors and constructing anchor graphs in a unified framework. \cite{10440580} constructs a consistent anchor graph and divides the anchor graph into a specific portion as well as a noise portion, and then imposes low-rank constraints on the specific portion. In addition, \cite{9146384} proposes an alternative anchor sampling strategy to construct individual anchor graphs, which are then merged into the consensus graph.

Although these methods have achieved considerable performance, the existing work still has the following considerations that can be improved: (1) anchor selection is performed independently in each view and lacks mutual consultation with other views, leading to poor performance of the learned representation. (2) Existing methods simply learn anchors dynamically but ignore the underlying cluster structure shared between anchors and raw data, resulting in poor-quality anchors. In the next section, this paper proposes a new MVC approach to address these issues.

\section{The Proposed Method}
In this section, the ALPC method proposed in this paper is described in detail and then an alternating optimization algorithm is provided to solve the resulting problem. This will be followed by a discussion of the time complexity as well as the convergence of ALPC.

\subsection{Formulation}
For anchors and graph learning, each view can be represented as a linear concatenation of all anchors. Based on this assumption, we obtain view-specific anchors and consensus subspace representation:

\begin{equation}
\begin{gathered}
\label{2}
\begin{gathered}
\begin{array}{l}
\mathop {\min }\limits_{{\mathbf{A}^{\left( v \right)}},z_i} \sum\limits_{v = 1}^l {\sum\limits_{i = 1}^n {\left\| {x_i^{\left( v \right)} - {\mathbf{A}^{\left( v \right)}}z_i} \right\|_2^2} } \\
s.t.{\mathbf{A}^{\left( v \right)T}}{\mathbf{A}^{\left( v \right)}} = \mathbf{I},z_i \ge 0,z_i^{T}\textbf{1} = 1.
\end{array}
\end{gathered} 
\end{gathered} 
\end{equation}
where $x_i^{\left( v \right)}$ and $z_i$ are the i-th columns of ${\mathbf{X}^{(v)}}$ and $\mathbf{Z}$, denoting the corresponding data and linear coefficients. Applying orthogonal constraints to ${\mathbf{A}^{(v)}}$ makes the learned anchors more diverse.

\begin{figure}[t] 
\centering 
\includegraphics[width=0.46\textwidth]{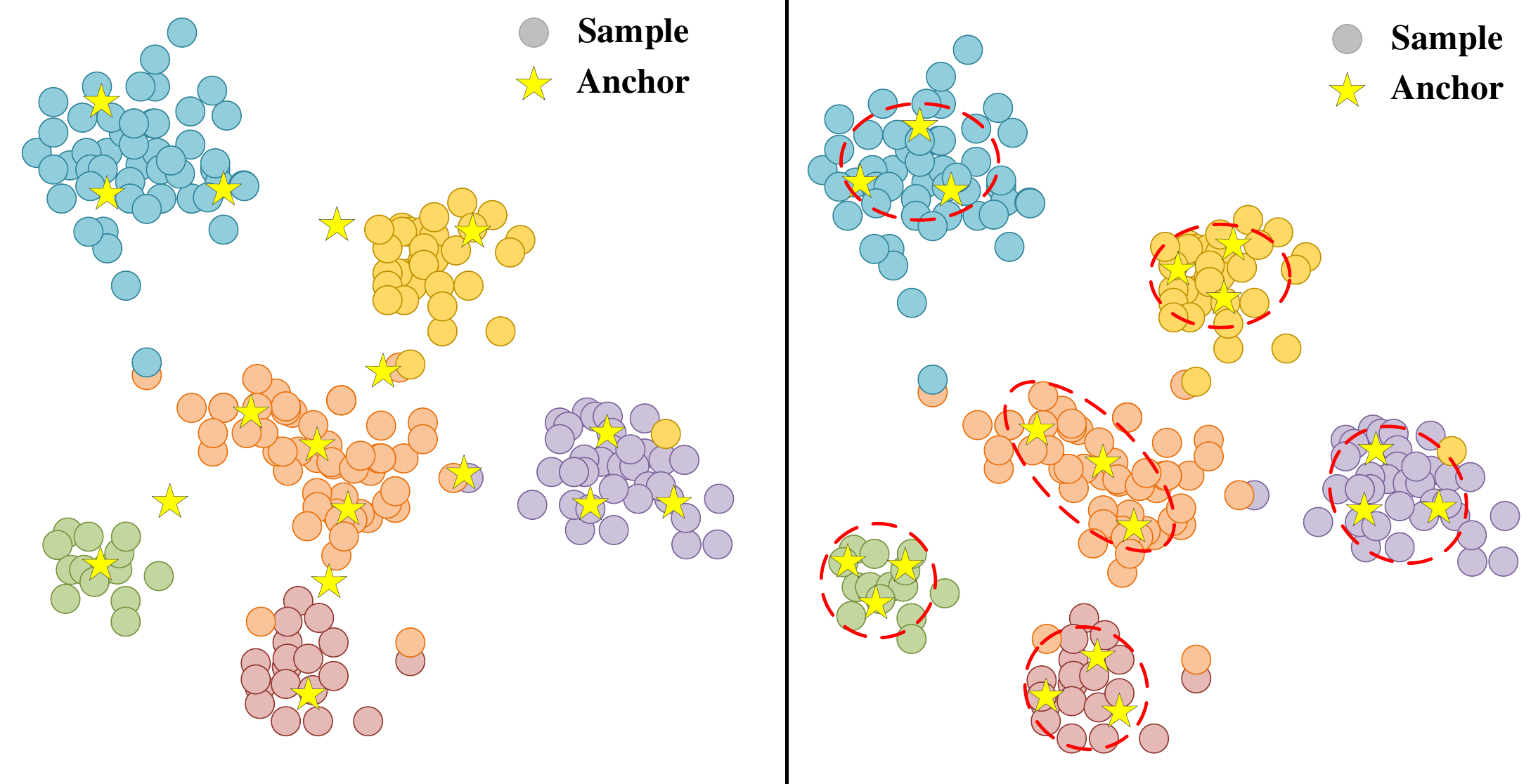} 
\caption{Traditional dynamic generation of anchors (left) and our proposed cluster-shared generation of anchors (right)}  
\label{fig1} 
\end{figure}

However, a simple orthogonality constraint does not guarantee that the generated anchors are representative, and such a constraint may result in some complex data not being represented. Fortunately, we can constrain the learning of anchors based on the real-world significance of anchors, i.e., representative anchors should be generated uniformly from different data clusters, as shown in Figure 2. 

Among them, in Figure 2, the left side is the traditional dynamic anchor learning strategy, and it can be seen that the anchors learned by the traditional method are not only unevenly distributed, some clusters contain a small number of anchors, and some clusters contain a large number of anchors, and some of the anchors are outside the clusters, and it is obvious that these outside anchors are not representative, because it may represent different cluster information. On the right is the new strategy of anchor learning proposed in this paper, i.e., generating anchors from clusters dynamically, so that the learned anchors are not only representative but also evenly distributed among different clusters. Inspired by this, we make the following constraints to learn more representative anchors:

\begin{equation}
\begin{gathered}
\label{3}
\begin{gathered}
\begin{array}{l}
\mathop {\min }\limits_\Phi  \sum\limits_{v = 1}^l {{\lambda _1}\left\| {{\mathbf{A}^{\left( v \right)}} - {\mathbf{U}^{\left( v \right)}}\mathbf{P}} \right\|_F^2 + {\lambda _2}\left\| {\mathbf{Z} - {\mathbf{P}^T}\mathbf{R}} \right\|_F^2} \\
{\rm{s}}.{\rm{t}}.{\mathbf{U}^{\left( v \right)T}}{\mathbf{U}^{\left( v \right)}} = \mathbf{I}{\rm{,}}{\mathbf{A}^{\left( v \right)T}}{\mathbf{A}^{\left( v \right)}} = \mathbf{I}
\end{array}
\end{gathered} 
\end{gathered} 
\end{equation}
where ${\mathbf{U}^{(v)}}$ is a basis matrix on which orthogonal constraint are imposed to avoid infinite solutions, ${\mathbf{P}}$ denotes the potential clustering indicators of the anchors, and ${\mathbf{R}}$ denotes the clustering indicators of the original data. In order to align the clustering center of mass of anchors with that of the original data, ALPC obtains the clustering center of mass of the original data as well as the clustering center of anchors by decomposing the anchor graph, thus constraining the two to share the same clustering structure and thus aligning the clustering centers of the two. Combining Eq. (2) and Eq. (3), we express the objective function of ALPC as follows:

\begin{equation}
\begin{gathered}
\label{4}
\begin{gathered}
\begin{array}{l}
\mathop {\min }\limits_\Phi  \sum\limits_{v = 1}^l {\sum\limits_i^n {\left\| {x_i^{\left( v \right)} - {\mathbf{A}^{\left( v \right)}}{z_i}} \right\|_2^2 + } } \\
\sum\limits_{v = 1}^l {{\lambda _1}\left\| {{\mathbf{A}^{\left( v \right)}} - {\mathbf{U}^{\left( v \right)}}\mathbf{P}} \right\|_F^2}  + {\lambda _2}\left\| {\mathbf{Z} - {\mathbf{P}^T}\mathbf{R}} \right\|_F^2\\
{\rm{s}}.{\rm{t}}.{\mathbf{U}^{\left( v \right)T}}{\mathbf{U}^{\left( v \right)}} = \mathbf{I}{\rm{,}}{\mathbf{A}^{\left( v \right)T}}{\mathbf{A}^{\left( v \right)}} = \mathbf{I},{z_i} \ge 0,z_i^T\textbf{1} = 1.
\end{array}
\end{gathered} 
\end{gathered} 
\end{equation}
where $\lambda _1$ and $\lambda _2$ are equilibrium parameters. Eq. (4) combines anchor graph learning and dynamic anchor learning into a unified framework. The first term is to learn anchors with diversity as well as anchor graphs, while the second term is to learn the underlying clustering structure of the anchors, and the third term is to align the anchors with the clustering centers of mass of the original data so that the anchors can be generated from the clustered clusters.

\begin{algorithm}[!ht]
    \renewcommand{\algorithmicrequire}{\textbf{Input:}}
	\renewcommand{\algorithmicensure}{\textbf{Output:}}
	\caption{\textbf{ALPC} algorithm}
    \label{power}
    \begin{algorithmic}[1] 
        \REQUIRE  Multi-view data $\{ {{\mathbf X^{( v )}}} \}_{v = 1}^l$, cluster number $k$, parameters ${\lambda _1}$, ${\lambda _2}$, $m$; 
   \renewcommand{\algorithmicrequire}{ \textbf{Initialize:}}
   \REQUIRE  
         ${\mathbf{A}^{( v )}}{\rm{ = 0}}$, ${\mathbf{U}^{( v )}}{\rm{ = 0}}$, $\mathbf{P} = 0$, $\mathbf{R} = 0$
	    \ENSURE Perform k-means on $\mathbf{Z}$ to obtain the clusters.
        \WHILE {not converged}
            \STATE  Update variable ${{\mathbf A^{( v )}}}$ using Eq. (7);
            \STATE  Update variable ${{\mathbf Z}}$ using Eq. (9);
            \STATE  Update variable ${{\mathbf P }}$ using Eq. (11);
            \STATE  Update variable ${{\mathbf R }}$ using Eq. (13);
            \STATE  Update variable ${{\mathbf U^{( v )}}}$ using Eq. (15);
        \ENDWHILE
    \end{algorithmic}
\end{algorithm}
\subsection{Optimization}
The optimization problem in Eq. (4) is not jointly convex when considering all variables simultaneously. Hence, we propose an alternating algorithm to optimize each variable
while the others maintain fixed.

$\textbf{Update\ $\mathbf{A}^{\left( v \right)}$}$ Since different ${\mathbf{A}^{\left( v \right)}}$ is independent of each other on different views, it is possible to solve each ${\mathbf{A}^{\left( v \right)}}$ independently. Fixing the other variables, the subproblem w.r.t. ${\mathbf{A}^{\left( v \right)}}$ is:
\begin{equation}
\begin{gathered}
\label{5}
\begin{gathered}
\begin{array}{l}
\mathop {\min }\limits_{\mathbf{A}^{\left( v \right)}} \sum\limits_{i = 1}^n {\left\| {x_i^{\left( v \right)} - {\mathbf{A}^{\left( v \right)}}{z_i}} \right\|_2^2 + } {\lambda _1}\left\| {{\mathbf{A}^{\left( v \right)}} - {\mathbf{U}^{\left( v \right)}}\mathbf{P}} \right\|_F^2\\
s.t.{\mathbf{A}^{\left( v \right)T}}{\mathbf{A}^{\left( v \right)}} = \mathbf{I}
\end{array}
\end{gathered} 
\end{gathered} 
\end{equation}
Setting the derivative of ${\mathbf{A}^{\left( v \right)}}$  to zero solves the above problem:
\begin{equation}
\begin{gathered}
\label{6}
\begin{gathered}
{\mathbf{A}^{\left( v \right)}}\mathbf{Z}{\mathbf{Z}^T} + {\lambda _1}{\mathbf{A}^{\left( v \right)}} - {\lambda _1}{\mathbf{U}^{\left( v \right)}}\mathbf{P} - {\mathbf{X}^{\left( v \right)}}{\mathbf{Z}^T} = \mathbf{0}
\end{gathered} 
\end{gathered} 
\end{equation}
Obviously, the optimal solution is:
\begin{equation}
\begin{gathered}
\label{7}
{\mathbf{A}^{\left( v \right)}} = \left( {{\lambda _1}{\mathbf{U}^{\left( v \right)}}\mathbf{P} + {\mathbf{X}^{\left( v \right)}}{\mathbf{Z}^T}} \right)\left( {\mathbf{Z}{\mathbf{Z}^T} + {\lambda _1}\mathbf{I}} \right)^{-1}
\end{gathered} 
\end{equation}

$\textbf{Update\ $\mathbf{Z}$}$ With fixing other variables, the sub-problem w.r.t. ${\mathbf{Z}}$ is:
\begin{equation}
\begin{gathered}
\label{8}
\mathop {\min }\limits_{\mathbf{Z}} \sum\limits_{v = 1}^l {\sum\limits_{i = 1}^n {\left\| {x_i^{\left( v \right)} - {\mathbf{A}^{\left( v \right)}}{z_i}} \right\|_2^2} }  + {\lambda _2}\left\| {\mathbf{Z} - {\mathbf{P}^T}\mathbf{R}} \right\|_F^2
\end{gathered} 
\end{equation}
Similar to solving $\mathbf{A}^{\left( v \right)}$, the optimal solution can be obtained by setting the derivative of $\mathbf{Z}$ to zeros:
\begin{small}
\begin{equation}
\begin{gathered}
\label{9}
\mathbf{Z} = {\left( {\sum\limits_{v = 1}^l {{\mathbf{A}^{\left( v \right)T}}{\mathbf{A}^{\left( v \right)}} + {\lambda _2}\mathbf{I}} } \right)^{ - 1}}\left( {\sum\limits_{v = 1}^l {{\mathbf{X}^{\left( v \right)T}}{\mathbf{X}^{\left( v \right)}} + {\lambda _2}{\mathbf{P}^T}\mathbf{R}} } \right)
\end{gathered} 
\end{equation}
\end{small}

$\textbf{Update\ $\mathbf{P}$}$ Fixing other variables optimization problem w.r.t. ${\mathbf{P}}$ is:
\begin{equation}
\begin{gathered}
\label{10}
\mathop {\min }\limits_{\mathbf{P}} 
\sum\limits_{v = 1}^l {\left( {{\lambda _1}\left\| {{{\mathbf{A}}^{(v)}} - {{\mathbf{U}}^{(v)}}{\mathbf{P}}} \right\|_F^2} \right) + {\lambda _2}\left\| {{\mathbf{Z}} - {{\mathbf{P}}^{{T}}}{\mathbf{R}}} \right\|_F^2} 
\end{gathered} 
\end{equation}
This can be solved by solving the following problem:
\begin{equation}
\begin{gathered}
\label{11}
{\mathbf{P}} = {\left( {{\lambda _1}{\mathbf{I + }}{\lambda _2}{{\mathbf{R}}^{{T}}}{\mathbf{R}}} \right)^{ - 1}}\left( {{\lambda _1}\sum\limits_{v = 1}^l {{{\mathbf{U}}^{(v)T}}{{\mathbf{A}}^{(v)}} + {\lambda _2}{\mathbf{R}}{{\mathbf{Z}}^T}} } \right)
\end{gathered} 
\end{equation}

$\textbf{Update\ $\mathbf{R}$}$ Fixing other variables optimization problem w.r.t. ${\mathbf{R}}$ is:
\begin{equation}
\begin{gathered}
\label{12}
\mathop {\min }\limits_\mathbf{R} {\lambda _2}\left\| {{\mathbf{Z} } - {{\mathbf{P} }^{{T}}}{\mathbf{R} }} \right\|_F^2
\end{gathered} 
\end{equation}

This can be solved by solving the following problem:

\begin{equation}
\begin{gathered}
\label{13}
{\mathbf{R}} = {\left( {{\mathbf{P}}{{\mathbf{P}}^{{T}}}} \right)^{ - 1}}{\mathbf{PZ}}
\end{gathered} 
\end{equation}

$\textbf{Update\ $\mathbf{U}^{(v)}$}$ Fixing other variables optimization problem w.r.t. $\mathbf{U}^{(v)}$ is:

\begin{equation}
\begin{gathered}
\label{14}
\mathop {\min }\limits_{{\mathbf{U}^{\left( v \right)}}} \left\| {{{\mathbf{A}}^{(v)}} - {{\mathbf{U}}^{(v)}}{\mathbf{P}}} \right\|_F^2 \quad s.t.{\mathbf{U}^{\left( v \right)T}}{\mathbf{U}^{\left( v \right)}} = \mathbf{I}
\end{gathered} 
\end{equation}
It is equivalent to solving:
\begin{equation}
\begin{gathered}
\label{15}
\mathop {\max }\limits_{{\mathbf{U}^{\left( v \right)}}} 
\mathbf{Tr}\left( {{{\mathbf{U}}^{(v)}}{{\mathbf{A}}^{(v)}}{{\mathbf{P}}^T}} \right) \quad s.t.{\mathbf{U}^{\left( v \right)T}}{\mathbf{U}^{\left( v \right)}} = \mathbf{I}
\end{gathered} 
\end{equation}
Since this is an orthogonal problem, it can be computed by singular value decomposition (SVD) of $\mathbf{A}^{(v)}\mathbf{P}^T$,i.e., ${\mathbf{A}^{\left( v \right)}}{\mathbf{P}^T} = \mathbf{U\Sigma {V^T}}$. The optimal solution is ${\mathbf{U}^{\left( v \right)}} = \mathbf{U{V^T}}$.

From the above optimization, it can be seen that ALPC needs to update five variables, and each subproblem has a corresponding solution. For convenience, the whole optimization process is summarized as Algorithm 1.

\subsection{Complexity and Convergence Analysis}
In this section, we will analyze the computational complexity of ALPC. It includes space complexity, time complexity and convergence.

\textbf{Memory complexity:} ALPC needs to store the matrices ${{\mathbf{Z}} \in {\mathbb{R}^{{mc} \times n}}}$, ${{\mathbf{A}^{(v)}} \in {\mathbb{R}^{{d}_v \times mc}}}$, ${{\mathbf{P}} \in {\mathbb{R}^{{c} \times mc}}}$. So the main space cost of ALPC is ${\cal O}\left( {ndv + nmv} \right)$, which proves its linear space complexity.
\begin{table*}[htbp!]
\centering
\renewcommand{\arraystretch}{1.25}
\scalebox{0.8451}{
\setlength{\tabcolsep}{7pt}
\begin{tabular}{|c|c|c|c|c|c|c|c|c|c|c|c|}
\hline
Dataset &  Metrics & LMVSC &  SMVSC & OPMC &  OMSC & SFMC & FPMVS & EOMSC & AWMVC & FDACF &  Ours\\
\hline
\multicolumn{1}{|c|}{\multirow{4}* {MSRC}}& ACC& 0.3476 & 0.6952 & 0.6286 & 0.7048 & 0.7286 & 0.6047 & 0.5905 & \underline{0.7810} & 0.6429  & \textbf{0.8819}\\
\multicolumn{1}{|c|}{} & NMI &  0.2461 & 0.6190 & 0.5533 & 0.6314 & 0.6987 & 0.5557 & 0.4352 & \underline{0.7160} & 0.6270 & \textbf{0.7758} \\
\multicolumn{1}{|c|}{} & Purity & 0.4048 & 0.6952 & 0.6381 & 0.7381 & 0.7667 & 0.6190 & 0.6048 & \underline{0.7810} & 0.7571 & \textbf{0.8819} \\
\multicolumn{1}{|c|}{} & F-score & 0.2480 & 0.5766 & 0.5125 & 0.6123 & 0.6736 & 0.5029 & 0.4261 & \underline{0.6852} & 0.5537 & \textbf{0.7637} \\
\hline
\multicolumn{1}{|c|}{\multirow{4}* {BBCSport}}& ACC & 0.6544 & 0.5091 & 0.6691 & 0.4522 & \underline{0.8254} & 0.4209 & 0.4154 & 0.6397 & 0.4669 & \textbf{0.9512}\\
\multicolumn{1}{|c|}{} & NMI &   0.4471 & 0.2118 & 0.6705 & 0.2129 & \underline{0.7770} & 0.1508 & 0.2367 & 0.4820 & 0.3749 & \textbf{0.8410} \\
\multicolumn{1}{|c|}{} & Purity & 0.6636 & 0.5367 & 0.7445 & 0.5110 & 0.8566 & 0.5183 & 0.5037 & 0.6890 & \underline{0.9228} & \textbf{0.9512}  \\
\multicolumn{1}{|c|}{} & F-score & 0.4854 & 0.3645 & 0.6501 & 0.3556 & \underline{0.7829} & 0.3247 & 0.3468 & 0.5234 & {0.4978} & \textbf{0.9009}  \\
\hline
\multicolumn{1}{|c|}{\multirow{4}* {Wiki}}& ACC&0.1884 & 0.3273 & 0.1766 & 0.3535 & 0.3039 & 0.3140 & \underline{0.5415} & 0.2170  & 0.5265 & \textbf{0.6040}\\
\multicolumn{1}{|c|}{} & NMI &   0.0469 & 0.1628 & 0.0454 & 0.1991 & 0.3196 & 0.1715 & \underline{0.5290} & 0.0783 & 0.5149 & \textbf{0.5514} \\
\multicolumn{1}{|c|}{} & Purity & 0.2083 & 0.3433 & 0.2069 & 0.3765 & 0.3458 & 0.3367 & 0.6141 & 0.2547 & \textbf{0.6811} & \underline{0.6399}  \\
\multicolumn{1}{|c|}{} & F-score & 0.1294 & 0.2103 & 0.1244 & 0.2243 & 0.2113 & 0.2146 & \underline{0.4832} & 0.1432 & {0.4325} & \textbf{0.5105}  \\
\hline
\multicolumn{1}{|c|}{\multirow{4}* {Caltech101-all}}& ACC&0.2005 & 0.2750 & 0.2188 & 0.2512 & 0.1777 & \underline{0.3015} & 0.2414 & 0.2367  & 0.2997 & \textbf{0.3284}\\
\multicolumn{1}{|c|}{} & NMI &  0.2145 & \underline{0.3510} & 0.2611 & 0.2522 & 0.2613 & 0.3549 & 0.2154 & 0.2397 & 0.2679 & \textbf{0.3697} \\
\multicolumn{1}{|c|}{} & Purity & 0.2396 & 0.3395 & 0.2691 & 0.2512 & 0.2854 & \underline{0.3460} & 0.2460 & 0.2691 & 0.2619 & \textbf{0.3617}  \\
\multicolumn{1}{|c|}{} & F-score & 0.1586 & 0.1462 & 0.0691 & 0.1422 & 0.0462 & \underline{0.2326} & 0.1154 & 0.0697 & 0.1669 & \textbf{0.2431}  \\
\hline
\multicolumn{1}{|c|}{\multirow{4}* {MNIST}}& ACC&0.9896 & 0.9884 & 0.8392 & 0.8922 & \underline{0.9905} & 0.9884 & 0.9808 & 0.9885  &0.9887 & \textbf{1.0000}\\
\multicolumn{1}{|c|}{} & NMI & 0.9685 & 0.9650 & 0.9134 & 0.9336 & \underline{0.9709} & 0.9650 & 0.9475 & 0.9647 & 0.9658 & \textbf{0.9879} \\
\multicolumn{1}{|c|}{} & Purity & \underline{0.9896} & 0.9884 & 0.9385 & 0.8922 & 0.9813 & 0.9768 & 0.9621 & 0.9761 & 0.9735 & \textbf{1.0000}  \\
\multicolumn{1}{|c|}{} & F-score & 0.9794 & 0.9768 & 0.8550 & 0.8923 & \underline{0.9813} & {0.9768} & 0.9621 & 0.9671 & 0.9735 & \textbf{0.9981}   \\
\hline
\multicolumn{1}{|c|}{\multirow{4}* {YouTubeFace\_sel}}& ACC& 0.1479 & 0.1749 & 0.1556 & 0.1897 & 0.2104 & \underline{0.2414} & 0.1924 & 0.2367 & 0.2146 & \textbf{0.3046} \\
\multicolumn{1}{|c|}{} & NMI & 0.1327 & 0.1091 & 0.0961 & 0.1225 & 0.0593 & 0.1327 & 0.1154 & \underline{0.2097} & 0.1966 & \textbf{0.2536}  \\
\multicolumn{1}{|c|}{} & Purity & 0.1216 & 0.1433 & 0.1069 & 0.1365 & 0.0958 & \underline{0.3279} & 0.1141 & 0.1547 & 0.0811 & \textbf{0.3299}   \\
\multicolumn{1}{|c|}{} & F-score & 0.0849 & 0.0903 & 0.0244 & 0.1243 & 0.0713 & \underline{0.1433} & 0.0832 & 0.1032 & 0.0732 & \textbf{0.1532}   \\
\hline
\end{tabular}}
\caption{ The Clustering performance of the compared multi-view clustering methods, including ACC, NMI, Purity and F-score. the best of these results are highlighted in bold. Underlining indicates sub-optimal performance.}
\label{Table3}
\centering
\end{table*}
\textbf{Time complexity:} The time consumption consists mainly of the update of five variables. For the update of the variable $\mathbf{Z}$, most of the time consumed is to find the product of the matrices and the inverse of the matrices, with a time complexity of ${\cal O}\left( {{m^3}{c^3} + {m^2}{c^2}d + {m^2}{c^2}n + mcdn} \right)$. Optimize the time complexity of $\mathbf{A}^{(v)}$ to ${\cal O}\left( {{m^3}{c^3} + {d_v}cmn} \right)$. For updating $\mathbf{P}$, then the complexity of ${\cal O}\left( {n{c^2}{d^2} + {c^3}{d^3} + 2nmcd} \right)$ is required. For updating $\mathbf{R}$, then the complexity of ${\cal O}\left( {{c^3}md} \right)$ is required. For updating $\mathbf{U}^{(v)}$, then the complexity of ${\cal O}\left( {{d_v}m{c^2}} \right)$ is required. For updating $\mathbf{P}$, then the complexity of ${\cal O}\left( {n{c^2}{d^2} + {c^3}{d^3} + 2nmcd} \right)$ is required. The computational complexity of the proposed ALPC is linearly related to the number of samples $n$ due to $m,d,v \ll n$. So overall time complexity is ${\cal O}\left( n \right)$.

\textbf{Convergence:} From Eq. (4), it can be seen that the whole function is not convex when all variables are considered simultaneously. So we use an alternating optimization algorithm to update the resolution for each variable. 

Let ${\cal J}\left( {{\mathbf{Z}_t},{\mathbf{A}_t},{\mathbf{P}_t},\mathbf{R}_t,\mathbf{U}_t} \right)$ denote the value of the objective function at the t-th iteration, it can be obtained as ${\cal J}\left( {\mathbf{Z}_t,\mathbf{A}_t,\mathbf{P}_t,\mathbf{R}_t,\mathbf{U}_t} \right) \le {\cal J}\left( {{\mathbf{Z}_{t + 1}},{\mathbf{A}_t},{\mathbf{P}_t},{\mathbf{R}_t},{\mathbf{U}_t}} \right) \le ... \le {\cal J}\left( {{\mathbf{Z}_{t + 1}},{\mathbf{A}_{t + 1}},{\mathbf{P}_{t + 1}},{\mathbf{R}_{t + 1}},{\mathbf{U}_{t + 1}}} \right)$. This shows that the objective function value of ALPC decreases monotonically. Since the lower bound value of the algorithm is zero, it can be obtained that the algorithm can converge to a minimum value.
\section{Experiment}
In this section, we conduct extensive experiments to compare ALPC with state-of-the-art methods and analyze the performance of the algorithm.
\subsection{Experiment Settings}
We validate the well-behaved performance of ALPC using six benchmark datasets, where the maximum number of samples used is more than 100,000 The datasets used in the experiments include \textbf{MSRC} \cite{CHEN20218}, $\textbf{BBCSprot}$\footnote{http://mlg.ucd.ie/datasets/bbc.html}, $\textbf{Wiki}$\footnote{http://www.svcl.ucsd.edu/projects/crossmodal/}, $\textbf{Caltech101-all}$\footnote{https://paperswithcode.com/dataset/caltech-101}, $\textbf{MNIST}$\footnote{https://yann.lecun.com/exdb/mnist/} and $\textbf{YouTubeFace sel}$\footnote{http://www.cs.tau.ac.il/~wolf/ytfaces/}. The details of the different datasets, including the number of samples, the number of views, the number of categories, and the feature dimensions of the dataset, are labeled in Table 3.

\begin{table}[!htbp]
\label{tab:tableTab} 
\scalebox{0.87}{
\begin{tabular}{c|cccc} 
\toprule 
Dataset & \hspace{0.1mm} $n$ & $v$ & $c$ & \quad $d$\\
\midrule 
$\mathbf{MSRC}$       & 210 &4 &7 &24/512/256/254\\
$\mathbf{BBCSport}$       & 544 &2 &5 &3183/3203\\
$\mathbf{Wiki}$       & 2866 &2 &10 &128/10\\
$\mathbf{Caltech101-all}$       & 9144 &5 &102 &48/40/254/512/928\\
$\mathbf{MNIST}$       & 60000 &3 &10 &342/1024/64\\
$\mathbf{YouTubeFace\_sel}$       & 101499 &5 &31 &64/512/64/647/838\\
\bottomrule 
\end{tabular}}
\raggedright
\caption{Summary of the used MVC benchmark datasets, where $n$, $v$, $c$, $d$ denote the number of samples, number of views, number of clusters, and feature dimensions, respectively} 
\end{table}
Our proposed algorithm is compared with nine state-of-the-art methods and the comparative algorithms are summarized as follows: $\textbf{LMVSC}$ \cite{Kang2019LargescaleMS}, $\textbf{SMVSC}$ \cite{10.1145/3474085.3475516}, $\textbf{OPMC}$ \cite{9710821}, $\textbf{OMSC}$\cite{10.1145/3534678.3539282}, $\textbf{SFMC}$\cite{9146384}, $\textbf{FPMVS}$\cite{9646486}, 
$\textbf{EOMSC}$\cite{Liu2022EfficientOM}, $\textbf{AWMVC}$\cite{Wan2023AutoweightedMC} and $\textbf{FDAGF}$\cite{Zhang2023LetTD}. Among them, the dynamic anchor learning strategies include SMVSC, OPMC, OMSC, FPMVS, EOMSC, AWMVC, FDAGF, and our proposed ALPC, while the static anchor strategies include LMVSC as well as SFMC. In this section, the clustering accuracies of the two anchor learning strategies will be compared, as well as the performance of ALPC in many dynamic anchor learning. For a fair comparison, we used the official codes of the corresponding methods and ran K-means 50 times to get the best results. The optimal parameters for these methods are the optimal parameters tuned by searching through the grid in the suggested range. For the methods in this paper, we adjust the values of $\lambda_1$ and $\lambda_2$ in $\left\{ {10^{-2},10^{-1},...10^2} \right\}$, $\left\{ {10^{-4},...,10^{-1},1} \right\}$. In this paper, four popular evaluation metrics are taken including Accuracy (ACC), Normalized Mutual Information (NMI), Purity and F-score. For all the metrics, higher values indicate superior method performance.
\subsection{Clustering Results}
Our comparison of ALPC and state-of-the-art methods was performed on six benchmark datasets on four metrics. Among them, Table 3 presents the results for ACC, NMI, Purity and F-score. From Table 3, the following conclusions are clearly observed:
\begin{itemize}
\item ALPC outperforms all other methods on most datasets. For example, the ACC of ALPC on MSRC, BBCsport, and MNIST are 10.09\%, 12.58\%, and 0.95\% higher than the second-best method, respectively, which demonstrates the effectiveness of our method.
\item Compared to LMVSC, better performance can be obtained by methods such as SMVSC, FPMVS, EOMSC and our proposed ALPC. The reason is that the former uses a static anchor selection strategy, while the latter all dynamically learn anchors and obtain an anchor graph, which makes the learned anchors more representative and discriminative, resulting in better clustering performance.
\item Most methods do so without considering that anchors should be uniformly distributed in each data cluster, and the quality of the learned anchors is poor. The learned anchor graphs are discriminative because ALPC introduces the shared potential consistency semantic constraints to guide the anchors to be generated uniformly and stably in each data cluster, which makes the anchors more representative and discriminative. This gives ALPC better and more stable clustering results.
\end{itemize}
\begin{figure}[htbp]
\vspace{-0.6cm}
	\centering
	\begin{subfigure}{0.48\linewidth}
		\centering
		\includegraphics[width=1\linewidth]{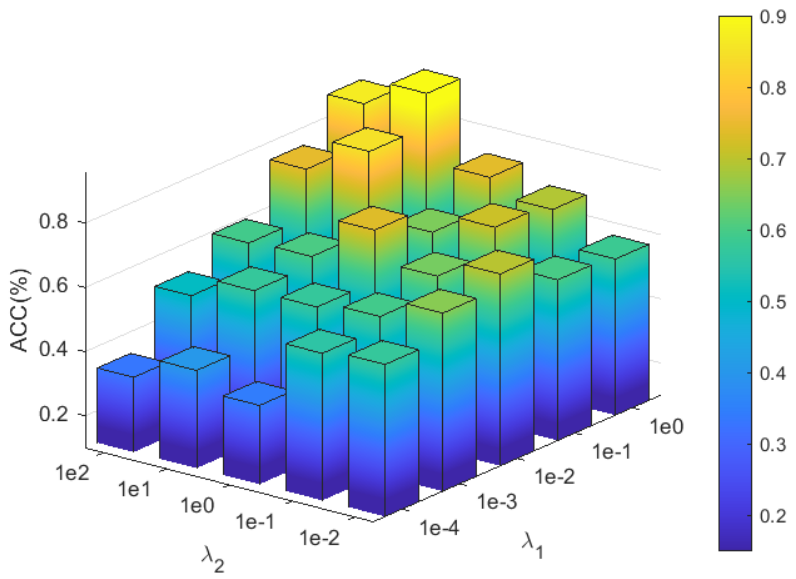}
		\caption{BBCSport}
		\label{BBCSport}
	\end{subfigure}
	\centering
	\begin{subfigure}{0.48\linewidth}
		\centering
		\includegraphics[width=1\linewidth]{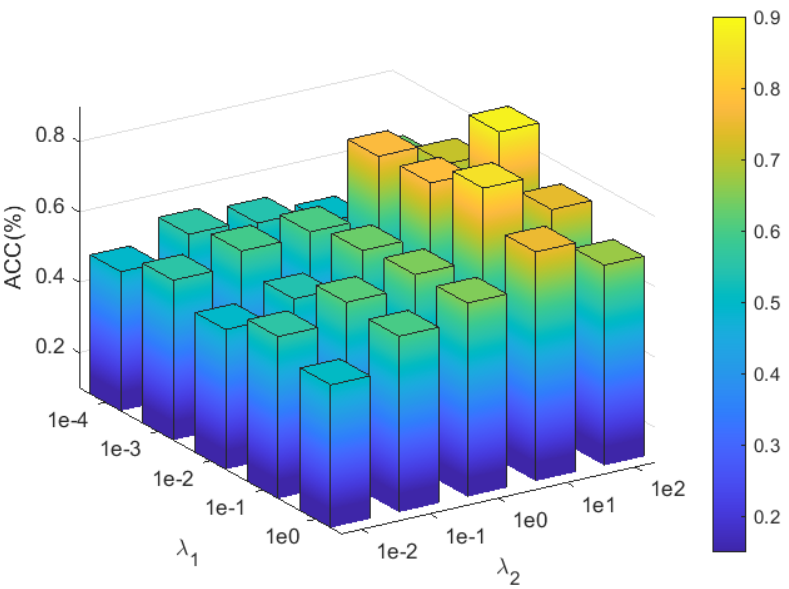}
		\caption{MSRC}
		\label{chutian3}
	\end{subfigure}

 	\centering
	\begin{subfigure}{0.48\linewidth}
		\centering
		\includegraphics[width=1\linewidth]{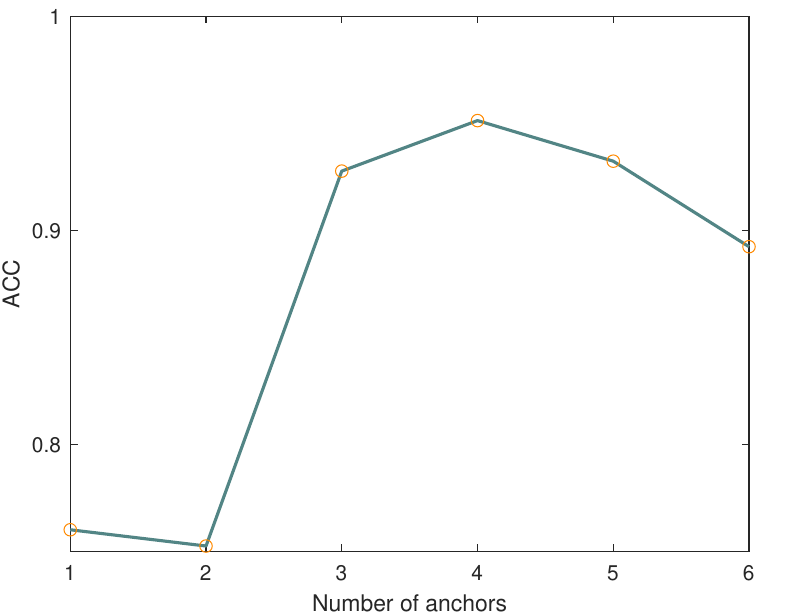}
		\caption{BBCSport}
		\label{BBCsport}
	\end{subfigure}
	\centering
	\begin{subfigure}{0.48\linewidth}
		\centering
		\includegraphics[width=1\linewidth]{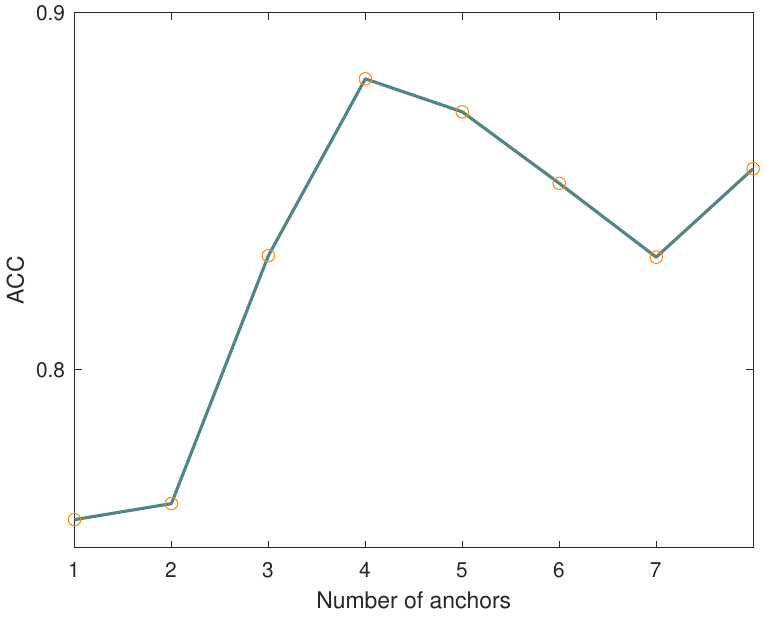}
		\caption{MSRC}
		\label{chutian3}
	\end{subfigure}
	\caption{(a) and (b) are sensitivity analyses of hyperparameters $\lambda_1$ and $\lambda_2$ to the test benchmark dataset, while (c) and (d) are sensitivity analyses of the number of anchors $m$ on two different datasets.}
	\label{da_chutian}
\end{figure}
\subsection{Parameter Analysis}
In this section, we use the grid search to tune two parameters $\lambda_1$ and $\lambda_2$ in ALPC in the ranges $\left\{ {10^{-4},...,10^{-1},1} \right\}$, $\left\{ {10^{-2},10^{-1},...10^2} \right\}$ thereby investigating the effect of the parameters on ALPC.

Among them, we conducted parameter sensitivity experiments on two benchmark datasets and recorded the clustering performance of ALPC for different combinations of the two parameters. Figure 3 shows the ACC values of ALPC for different combinations of parameters. From the figure 3, we can get the following conclusions: (1) When the values of parameters $\lambda_1$ and $\lambda_2$ are located at $\left\{ {10^{-1},1,10} \right\}$, ALPC can get the best performance. (2) When parameter $\lambda_2$ is too small, the results drop very fast, which indicates that ALPC is more sensitive to parameter $\lambda_2$. In addition, we also find that the number of anchors has a very large impact on the clustering performance. Obviously, most datasets perform best when the number of anchors is 2c or 3c. This proves the importance of anchor selection.

\subsection{Qualified Study}
To demonstrate the superiority of the consensus graph learned by ALPC, we visualized the consensus graph of APLC on two benchmark datasets, as shown in Figure 4. In the figure 4, we show the consensus graphs learned by APLC on $\mathbf{BBCSport}$ and $\mathbf{MSRC}$ dataset. A clear block structure corresponding to the clustering of a large number of samples can be observed in these two datasets, which indicates that the consensus graph learned by APLC has a well-defined block structure.

\begin{figure}[htbp]
	\centering
	\begin{subfigure}{0.48\linewidth}
		\centering
		\includegraphics[width=1\linewidth]{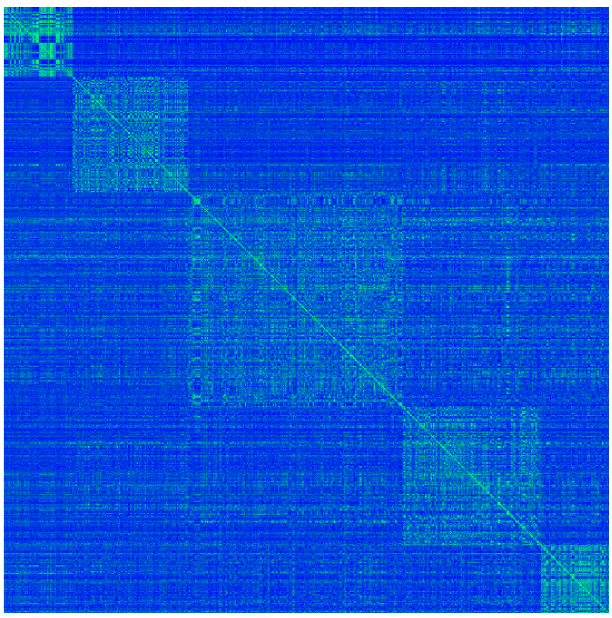}
		\caption{BBCsport}
		\label{BBCSport}
	\end{subfigure}
	\centering
	\begin{subfigure}{0.48\linewidth}
		\centering
		\includegraphics[width=1\linewidth]{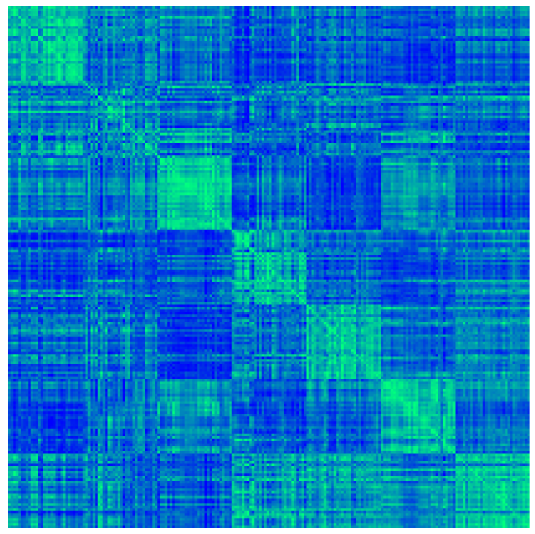}
		\caption{MSRC}
		\label{chutian3}
	\end{subfigure}
	\caption{The visualization of the complete graph of ALPC on both datasets BBCSport and MSRC.}
	\label{da_chutian}
\end{figure}

\subsection{Convergence and Time Comparison}
As mentioned earlier, the objective value of ALPC decreases monotonically with alternating updates of the variables and the objective function is lower bounded. In this section, we experimentally demonstrate the convergence of the ALPC algorithm, and the convergence curves for MNIST and Wiki data are given in Figure 5. It is clear from the figure that the value of the objective function decreases rapidly and stabilizes at an iteration number of about 20, thus the method is highly convergent.

\begin{figure}[htbp]
	\centering
	\begin{subfigure}{0.48\linewidth}
		\centering
		\includegraphics[width=1\linewidth]{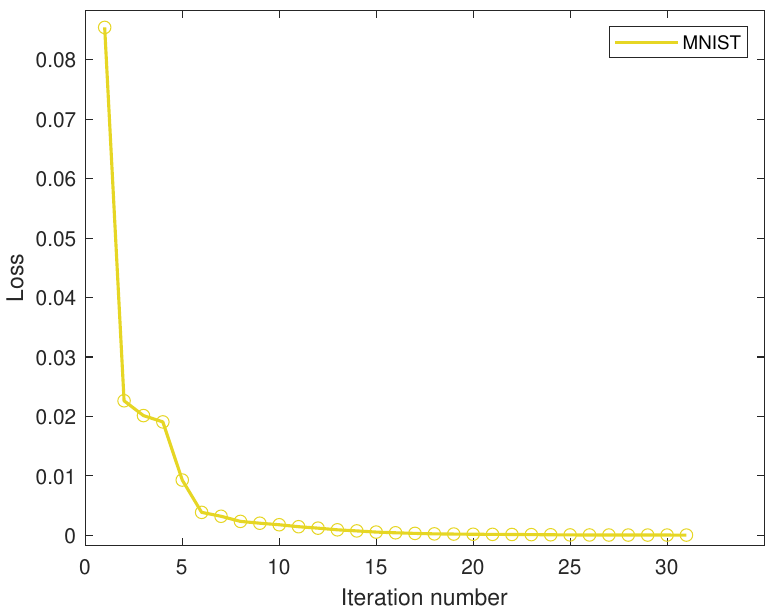}
		\caption{MNIST}
		\label{MNIST}
	\end{subfigure}
	\centering
	\begin{subfigure}{0.48\linewidth}
		\centering
		\includegraphics[width=1\linewidth]{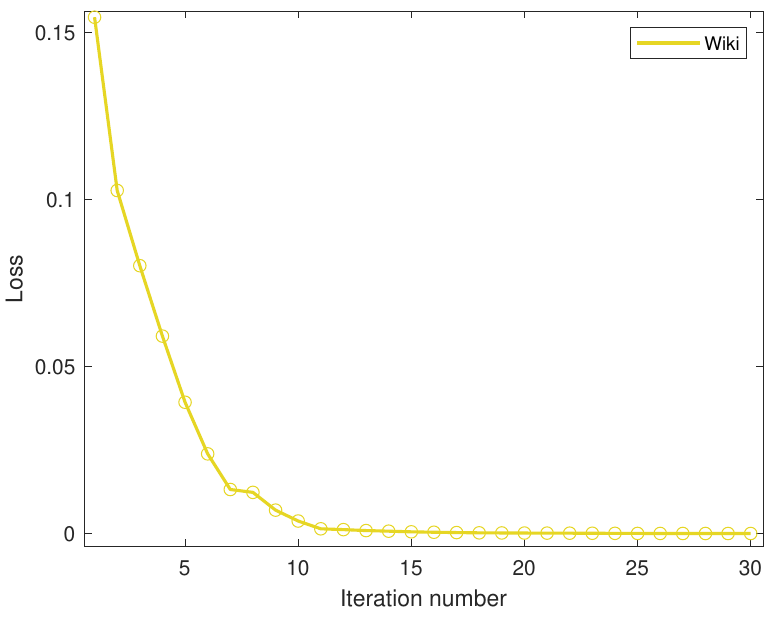}
		\caption{Wiki}
		\label{chutian3}
	\end{subfigure}
	\caption{ Convergence curves of ALPC on two datasets MNIST and Wiki.}
	\label{da_chutian}
\end{figure}

To be fair, we compared the computation time of ALPC and other methods on all datasets. The results are shown in Figure 6, where the y-axis is scaled by logarithms to mitigate the gap between the methods. It is clear from the figure 6 that ALPC is effective and takes clearly less computation time than the anchor-based MVC method, which proves that ALPC does not increase the computation time excessively.

\begin{figure}[htb] 
\centering 
\includegraphics[width=0.45\textwidth]{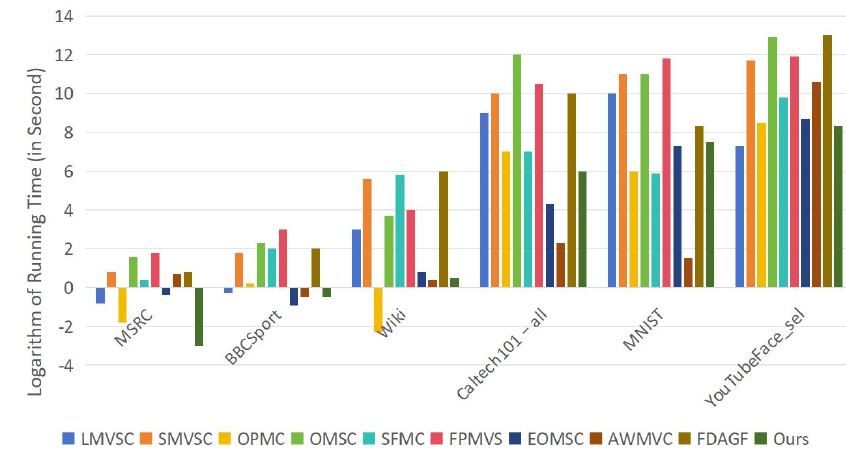} 
\caption{The relative running time of the compared algorithms and ALPC on the six benchmark datasets.}  
\label{fig1} 
\end{figure}

\subsection{Ablation Study}
While existing methods simply utilize the orthogonal constraint ${\mathbf{A}^{\left( v \right)T}}{\mathbf{A}^{\left( v \right)}} = \mathbf{I}$ for dynamic learning of anchors, ALPC takes into account the real-world significance of anchors. In this section, we conduct ablation experiments to evaluate the effectiveness of the proposed shared latent clustering structure. We utilize Eq. 2 and impose regularization constraints on the anchor graph, which has been widely used in previous methods. We compared this variant (Our-a) with our proposed ALPC method on all datasets and the results are shown in Table 4.
\begin{table}[htbp!]
\centering
\scalebox{0.822}{
\begin{tabular}{c|c|cccc}
\toprule
Dataset & Method & ACC & NMI & Purity & F-score \\
\midrule
{\multirow{2}* {MSRC}} & Ours-a & 0.7507 & 0.7334 & 0.7688 & 0.7683 \\
{\multirow{2}* {}} & Ours & \textbf{0.8819} & \textbf{0.7758} & \textbf{0.8819} & \textbf{0.7637} \\
\midrule
{\multirow{2}* {BBCSport}} & Ours-a & 0.8956 &\textbf{ 0.8926} & 0.8987 & 0.8614 \\
{\multirow{2}* {}} & Ours & \textbf{0.9512} & {0.8410} & \textbf{0.9512} & \textbf{0.9009} \\
\midrule
{\multirow{2}* {Wiki}} & Ours-a & 0.5201 & 0.5254 & 0.5313 & \textbf{0.5342} \\
{\multirow{2}* {}} & Ours & \textbf{0.6040} & \textbf{0.5514} & \textbf{0.6399} & {0.5105} \\
\midrule
{\multirow{2}* {Caltech101-all}} & Ours-a & 0.1007 & 0.1034 & 0.0889 & 0.0983 \\
{\multirow{2}* {}} & Ours & \textbf{0.3284} & \textbf{0.3697} & \textbf{0.3617} & \textbf{0.2431} \\
\midrule
{\multirow{2}* {MNIST}} & Ours-a & 0.9108 & 0.9015 & 0.9026 & 0.9197 \\
{\multirow{2}* {}} & Ours & \textbf{1.0000} & \textbf{0.9879} & \textbf{1.0000} & \textbf{0.9981} \\
\midrule
{\multirow{2}* {YouTubeFace\_sel}} & Ours-a & 0.1530 & 0.0734 & 0.1688 & 0.0683 \\
{\multirow{2}* {}} & Ours & \textbf{0.3046} & \textbf{0.2536} & \textbf{0.3299} & \textbf{0.1532} \\
\bottomrule
\end{tabular}}
\caption{Comparison of the performance of Ours-a and Ours on six benchmark datasets.}
\label{Table5}
\centering
\vspace{-0.3cm}
\end{table}
From the table, it can be clearly observed that the clustering performance of ALPC is better than that of Ours-a in all the datasets, which suggests that sharing the underlying clustering structure helps in the generation of high-quality anchors, and the generated anchors share a consistent clustering structure with the original data, which results in the anchors learned by ALPC being more representative and discriminative. These reasons result in ALPC being able to learn higher quality anchor graphs, which leads to better clustering performance. From these ablation results, it is shown that sharing the underlying clustering structure is significant for the improvement of clustering performance.

\section{Conclusion}
In this paper, we propose a new Anchor Learning with Potential Cluster Constraints for Multi-view Clustering method (ALPC). ALPC integrates and optimizes anchor selection and anchor graph learning into a unified framework. Unlike previous approaches, ALPC imposes shared potential clustering semantic constraints along with orthogonal constraints on anchors. Aligning the clustering centers of data clusters with the clustering centers of anchors constrains anchors to be generated uniformly in data clusters, making anchors more representative as well as discriminative This allows for exploring inter-cluster diversity and intra-cluster consistency. ALPC truly realizes the relevance of anchors, and experiments demonstrate the effectiveness of the method.

\bibliography{aaai22.bib}

\end{document}